\documentclass[sigconf,authorversion]{acmart}

\usepackage{algpseudocode}
\usepackage{algorithm}
\usepackage[disable]{todonotes}
\usepackage{balance}

\usepackage{todonotes}
\usepackage{url}

\newcommand{\new}[1]{\textcolor{black}{#1}}

\newcommand{\medbert}{\textsc{MedBERT}}

\AtBeginDocument{%
  \providecommand\BibTeX{{%
    \normalfont B\kern-0.5em{\scshape i\kern-0.25em b}\kern-0.8em\TeX}}}

\copyrightyear{2021}
\acmYear{2021}
\setcopyright{acmlicensed}\acmConference[CIKM '21]{Proceedings of the 30th ACM International Conference on Information and Knowledge Management}{November  1--5, 2021}{Virtual Event, QLD, Australia}
\acmBooktitle{Proceedings of the 30th ACM International Conference on Information and Knowledge Management (CIKM '21), November  1--5, 2021, Virtual Event, QLD, Australia}
\acmPrice{15.00}
\acmDOI{10.1145/3459637.3482128}
\acmISBN{978-1-4503-8446-9/21/11}

\begin{document}
\fancyhead{}
\title{Knowledge-Aware Neural Networks for Medical Forum Question Classification}


\author{Soumyadeep Roy}
\authornote{Also affiliated to L3S Research Center, Germany}
\affiliation{
\institution{IIT Kharagpur \country{India}}}

\author{Sudip Chakraborty}
\affiliation{
\institution{IIT Kharagpur \country{India}}}

\author{Aishik Mandal}
\affiliation{
\institution{IIT Kharagpur \country{India}}}

\author{Gunjan Balde}
\affiliation{
\institution{IIT Kharagpur \country{India}}}

\author{Prakhar Sharma}
\affiliation{
\institution{IIT Kharagpur \country{India}}}

\author{Anandhavelu Natarajan}
\affiliation{\institution{Adobe Research \country{India}}
}

\author{Megha Khosla}
\affiliation{\institution{L3S Research Center \country{Germany}}}

\author{Shamik Sural}
\affiliation{
\institution{IIT Kharagpur \country{India}}}

\author{Niloy Ganguly}
\authornotemark[1]
\affiliation{
\institution{IIT Kharagpur \country{India}}}

\renewcommand{\shortauthors}{S. Roy et al.}

\begin{abstract}
  Online medical forums have become a predominant platform for answering health-related information needs of consumers. However, with a significant rise in the number of queries and the limited availability of experts, it is necessary to automatically classify medical queries based on a consumer's intention, so that these questions may be directed to the right set of medical experts. Here, we develop a novel medical knowledge-aware BERT-based model (\medbert) that explicitly gives more weightage to medical concept-bearing words, and utilize domain-specific side information obtained from a popular medical knowledge base. We also contribute a multi-label dataset for the Medical Forum Question Classification (MFQC) task. \medbert\ achieves state-of-the-art performance on two benchmark datasets and performs very well in low resource settings.
\end{abstract}



\maketitle

\section{Introduction}\label{sec:intro}
Online medical or health forums have become quite popular and presently act as a reliable source of health-related information, advice or support~\cite{Sinha2018}. Given the limited availability of medical professionals, researchers have developed automated systems to detect the quality of an online health article~\cite{Afsana2021} or to identify the health information need category of a user based on its medical forum question in order to direct it to the relevant medical expert ~\cite{Derksen2017,jalan2018medical}. In this paper, we focus on the \textit{Medical Forum Question Classification} (\textbf{MFQC}) task, where the aim is to classify a health forum question~(either a query or query plus description) based on the intended health information need category of a user posting it. For example, the target class \textit{Treatment} is concerned with a specific medical procedure like surgery, while \textit{Family Support} class corresponds to issues related to a caregiver (instead of a patient) like how to support one's spouse or an ill child~\cite{verma2016classification,jalan2018medical}. Currently, MFQC is formulated as a single-label multi-class prediction problem~\cite{jalan2018medical}. Existing works for medical question classification usually employ  hand-crafted features or pretrained embeddings~\cite{Liu2011consqa,Kirk2016resourcemedqa,guo2017classifying,mrabet2016combining,verma2016classification} as input representations which result in the loss of context information present in the raw input. Moreover, as these representations are constructed using limited and noisy data (because of large presence of contractions and misspellings as well as vocabulary mismatch with medical professionals), these do not generalize well for the test data. 

To overcome the above problems, first, we 
 utilize pretrained models to extract statistically powerful context preserving representations from the raw input data. Pretrained models like BERT~\cite{devlin2019bert}, GPT-2~\cite{radford2019language} and RoBERTa~\cite{liu2019roberta} are usually trained on a large number of documents in a self-supervised manner to obtain general-purpose language representations. Task specific fine-tuning~\cite{howard-ruder-2018-universal} is usually preferred over the feature extraction of pretrained embeddings~\cite{peters-etal-2018-deep} and has shown vast improvements in several natural language processing tasks~\cite{zhang-bowman-2018-language}. Second, we provide medical domain knowledge as side information to our model, by providing explicit importance to medical concept-bearing words or what we call \emph{aspects}. This in turn helps us to learn both contextualized as well as a medical domain knowledge-aware representation of a medical forum question. The first standalone, medical knowledge-aware neural model (SoA-DN) to use additional medical domain features (biomedical entities extracted using Metamap~\cite{aronson2001effective}) for the MFQC task, fails to achieve high classification performance~\cite{jalan2018medical}. In general, several works~\cite{weissenborn2017dynamic,mihaylov2018knowledgeable,annervaz2018learning,yang2017leveraging} investigated novel strategies to incorporate domain-specific side information present as structured data in knowledge bases or knowledge graphs into neural models (including transformer-based pretrained models). 
 
We make the following contributions in this paper. \textbf{First}, we propose a novel application of dual encoder model~\cite{Yang2019AML,zeng2019lcf,bhowmik-etal-2021-fast} for MFQC task. Our approach,
\medbert, employs BERT encoders  to extract global context as well as medical context aware representations of the documents. We additionally enhance our representations by incorporating additional domain-specific side information. In particular, we attribute explicit importance to medical concept bearing terms in our final representations by a simple masking procedure. \medbert\ outperforms the state-of-the-art models across on two benchmark datasets of ICHI and CADEC and particularly works well in low-resource setting (using 30\% training data). Prior studies~\cite{jalan2018medical} observe that the target classes of ICHI are mutually overlapping and thus multi-label classification will be a better formulation for MFQC task. However, to the best of our knowledge, there does not exist any multi-label medical forum text classification dataset, and thus rigorously annotate the CADEC dataset~\cite{KARIMI201573}, a benchmark adverse drug event medical forum dataset across four health information search categories (see Section~\ref{sec:expt-setup}); this forms our \textbf{second} contribution. \textbf{Finally}, we perform extensive analysis to better understand the workings of \medbert, and make the codebase publicly available at \url{https://github.com/roysoumya/knowledge-aware-med-classification}. 
\section{Methodology}\label{sec:method}

\begin{figure}[!ht]
\centering
\includegraphics[scale=0.25]{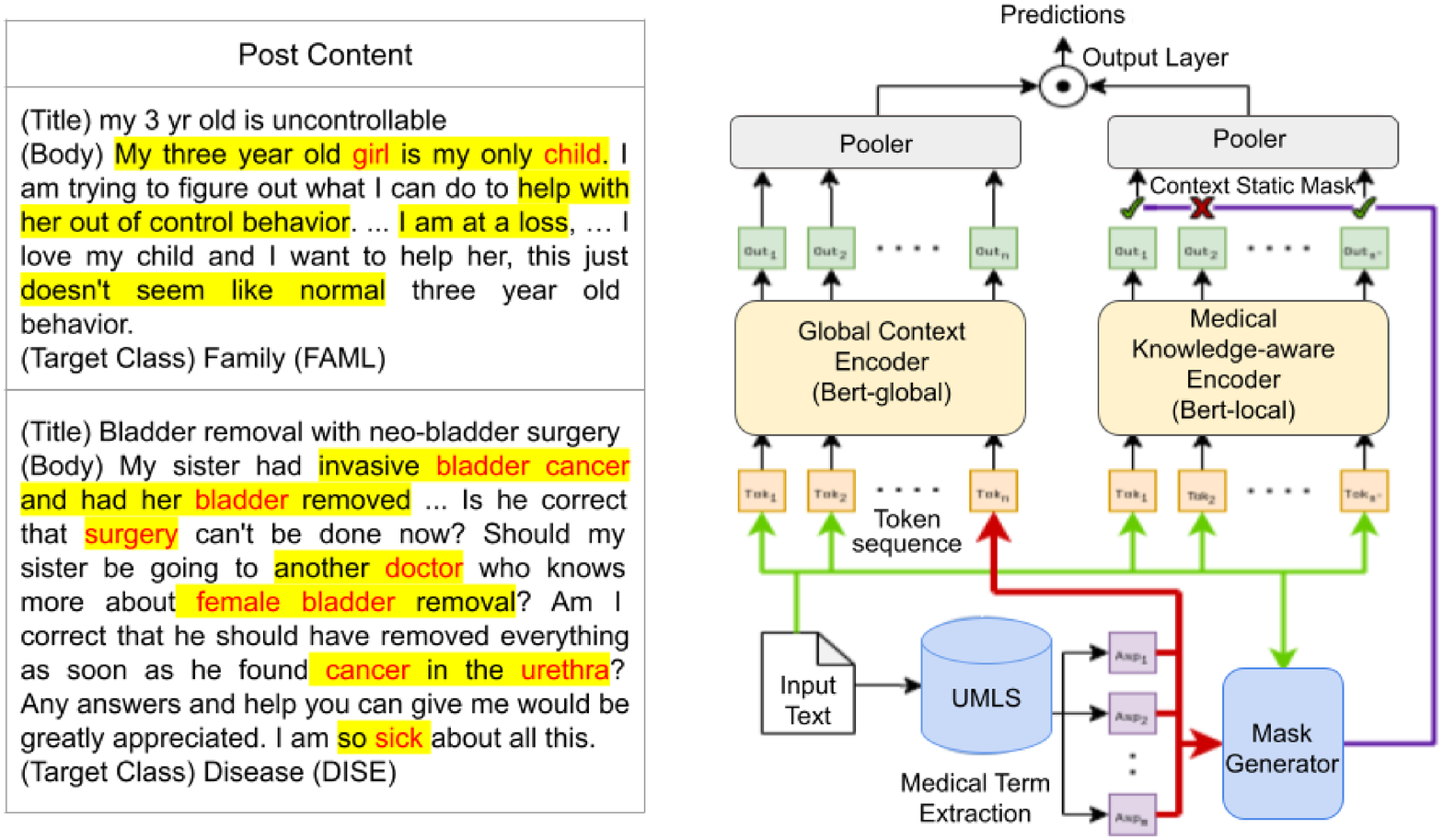}
\caption{(left) Anecdotal example from ICHI dataset where medical concept-bearing aspects (marked red) are highlighted along with context  (right) Methodology overview of \medbert\ for MFQC task}
\label{fig:ichi-framework-overview}
\end{figure}

\medbert\ is comprised of four modules (see Figure~\ref{fig:ichi-framework-overview} for methodology overview) --- (i) medical term extraction using a medical knowledge base, \new{which serve as domain-specific side information to our model}~(Section~\ref{sec:med-term-extract}), (ii) a BERT-based encoder (Bert\textsubscript{global}) that considers both medical words as well as their context (Section~\ref{sec:bert-global}), (iii) a BERT-based encoder (Bert\textsubscript{local}) that explicitly gives more importance to medical concept-bearing words~(Section~\ref{sec:med-knowledge-aware}), and (iv) merging the representations learnt by Bert\textsubscript{global} and Bert\textsubscript{local} (Section~\ref{sec:merge-global-local}) for final class prediction.

\subsection{Extraction of Medical Aspects }\label{sec:med-term-extract}
Here, we describe the identification of medical concept-bearing words in particular, and medically relevant words in general, from the input text. To put it more formally, as previously described, we represent the input text as a sequence of words, $x^{(i)} = { w_{1} , w_{2} , ... , w_{n} }$, 
we aim to obtain a subsequence from $x^{(i)}$, $x^{(i, t)} = w_{1}^t , w_{2}^t , ... , w_{m}^t$, where  $1 \leq m \leq n $. Here, $x^{(i, t)}$ represents the tokens that is (or a part of) a medical concept-bearing word (or phrase).

 The aspects are UMLS medical concepts, which are extracted using the QuickUMLS~\cite{soldaini2016quickumls} tool. However, we observe that not all extracted medical entities are meaningful for the task at hand. We, thus select only sixteen \textit{UMLS Semantic Classes}, as used by \citet{jalan2018medical}, that are closely associated with the target classes of MFQC task. As the posts are made by consumers not belonging from medical background, the medical vocabulary used is quite different from that contained in medical knowledge bases like UMLS. Thus we also utilize a medical lexicon known as \textit{Patient-Friendly Term List}~\cite{meddra-terms}, that contains around $1400$ MedDRA Lowest Level Terms which include the most frequently used vocabulary by patients or consumers for reporting adverse events. Some examples include \textit{Aching in limb, Blood pressure inadequately controlled, Crawling sensation of skin}. 
In the following sections we describe how we incorporate the importance of these terms to learn medical domain specific contextual representations.

\subsection{Global Context Representation (Bert\textsubscript{global})}\label{sec:bert-global}

To learn a global representation we follow the design of LCF~\cite{zeng2019lcf}. In particular we append the aspect terms to the original document sequence to serve as input to our global BERT encoder. 

\noindent \textbf{Input.}
 The global context sequence for a given medical forum question is constructed in following manner: \textit{input text + aspect\textsubscript{1} + aspect\textsubscript{2} + ... + aspect\textsubscript{n}} where the \textit{input text} and \textit{aspects} are separated by the special token [SEP]. The input to BERT encoder will be: \textit{[CLS]+ `I had surgery for retinal detachment in December.' + [SEP] + `surgery' + `retinal detachment' + [SEP] }.

\noindent \textbf{Text encoder (BERT) details.}
For the \medbert\ model, we use the pretrained BERT\textsubscript{base} model as proposed by \citet{devlin2019bert}. BERT is a multi-layer bidirectional Transformer encoder based on its original implementation as described by Vaswani et al.~\cite{Vaswani2017transformers}. Here, we use the \textit{uncased} variant which ignores the differences in case of a word (like cricket and Cricket). We apply the standard fine-tuning technique~\cite{devlin2019bert} where the final hidden vector corresponding to the  \textit{[CLS]} token is used as the aggregate representation for the baseline BERT model. 

\subsection{Medical Knowledge-Aware Representation (Bert\textsubscript{local})}\label{sec:med-knowledge-aware}

We now describe the steps involved in obtaining the medical domain knowledge-aware representation using BERT encoder, where (i) a deep representation of the  input text is generated using the BERT encoder, and then (ii) the deep document representation is modified using a \textit{context static mask} that forces BERT to prioritize medically-relevant words over the remaining words, thus making it a medical domain knowledge-aware representation.

\noindent \textbf{Input.}
 The input to the local context sequence consists of simply the sequence of tokens for a given document. We follow the standard method where the tokens are surrounded by \textit{[CLS]} and \textit{[SEP]} on the left and right, respectively. For example, if the input text is `I had surgery for retinal detachment in December.', the input to our encoder will be: \textit{[CLS]+ I had surgery for retinal detachment in December. + [SEP]  }.
 
\noindent \textbf{Learning aspect-aware representations}
Medical features are crucial for improving the performance of medical text-classification and using just the text in isolation fails to capture the required domain knowledge. We treat these features as aspects (marked red in the document shown in Figure~\ref{fig:ichi-framework-overview}) which our model should give more attention.
To learn aspect aware representations, we adopt a simple feature masking strategy as in \cite{zeng2019lcf}. In particular, we apply a context static mask over the output  representation of Bert\textsubscript{local} encoder such that the output representations corresponding to all words which are not aspects are masked or set to $0$.
These retained representations are both domain-specific and contextualized. In essence, the output tokens from the local context encoder are being filtered using masked entity representations. In the above example, for the sentence `I had surgery for retinal detachment in December', the medical aspects extracted are \textit{surgery} and \textit{retinal detachment}. Assuming, each token belongs within the vocabulary, the corresponding context static mask will be a binary vector: $[00101100]$

\subsection{Class Prediction}
\label{sec:merge-global-local}

To obtain the final representations, we concatenate the local and global context representation vectors as: $\overline{o} = [BERT\textsubscript{local} ; BERT\textsubscript{global}]$
 It is then passed through a fully-connected layer to project the concatenated vector into the space of the targeted $L$ classes :
$\overline{o}_{dense} = \overline{W}^{T} \overline{o} + \overline{b}$. 
Finally the posterior class is obtained by applying softmax on $\overline{o}_{dense}$. We enforce weight sharing for both the local and global context encoders in \medbert. \medbert\ uses the cross-entropy loss function for single-label tasks and `binary cross-entropy' for multi-label classification tasks.
\section{Experimental Evaluation}\label{sec:experiments}
We evaluate our proposed \medbert\ model on benchmark dataset as well as create an annotated corpus of the multi-label dataset and also test on it.
 We run each model for five times and report the mean performance scores in Table~\ref{tab:ichi-perf-compare}. We perform the McNemar's test~\cite{mcnemar1947note} for significance testing.

\subsection{Experimental Setup}\label{sec:expt-setup}

\noindent \textbf{Datasets.} We evaluate on two document-level datasets --- 
\todo[inline]{make dataset description shorter and add baseline description}

\noindent \textbf{(i) ICHI~\cite{ichi2016}}: It is a single-label multi-class MFQC dataset, comprising of questions posted on health discussion forums. It consists of $8000$ training and $3000$ test data points. Each data point consists of a question title, question text and label (belonging to one of seven health information need categories like Demographic (DEMO), Disease (DISE), Treatment (TRMT), Goal-oriented (GOAL), Pregnancy (PREG), Family Support (FMLY), Socializing (SOCL)). The dataset is balanced and we use the standard train-test splits employed by prior studies~\cite{jalan2018medical,verma2016classification}. 

\noindent \textbf{(ii) CADEC}: It is the first multi-label MFQC dataset, comprising of $942$ training and 300 testing data points. We utilize a benchmark annotated corpus~\cite{KARIMI201573} consisting of pharmacovigilance-related consumer questions/reviews, and use two computer-science graduate students who are very familiar with online forums but limited exposure to online medical forums, to label the data points into four health information search categories \new{consisting of Uncertainty of post-diagnosis (UPD), Medical Assistance (MAS), Diet and maintenance (DM), and Information source (IS)} (appropriate for posts related to `adverse drug events'), following the same guidelines as \citet{Derksen2017}; we remove data points where two annotators disagree with each other. The dataset suffers from high class imbalance and thus during training, multi-class oversampling is used where minority classes are selected one-by-one and then oversampled to the cardinality of the initial majority class.

\noindent \textbf{Baselines.} We compare with three types of baseline models --- (i) non-neural models like TFIDF + SVM, (ii) standard neural models based on pretrained embeddings that do not use medical domain knowledge like FastText~\cite{joulin2017bag}, TextCNN~\cite{TextCNNKim14}, HAN~\cite{yang2016hierarchical} and ICHI 2016 Challenge Winners~\cite{verma2016classification} (iii) medical domain knowledge-aware neural models like SoA-DN and H-BiLSTM + TFIDF-DN + SoA-DN, model variants developed by \citet{jalan2018medical}; we are unable to reproduce their results and thus report their overall accuracy scores, the only metric reported in their paper, (iv) medical domain knowledge-aware transformer-based pretrained models like \new{Local Context Focus Bert} (LCF-BERT)~\cite{zeng2019lcf}, Bert-base~\cite{devlin2019bert}, \medbert\ variants (Global only, Local only required for ablation study purposes).

\noindent \textbf{Training Details.} Here, we use the BERT implementation of transformers~\cite{hugtransformers} library. For preprocessing BERT-based models, we perform lowercasing and set maximum sequence length to 256 tokens with longer documents truncated from the right. Whereas for non-BERT based models, each word is converted to lower case and punctuations are removed, but stopwords are retained; the words are also lemmatized, and maximum document length is set to $3000$ words.  We perform hyperparameter-tuning for MedBERT in terms of learning rate (5e-5, 3e-5, 2e-5), batch size ($16$ and $32$) and number of epochs for fine-tuning ($2$ to $10$ epochs). \medbert\ model used in our experiments is trained with a batch size of $16$, \textit{Adam} as optimizer with a starting learning rate of 5e-5, linearly decayed throughout training and dropout rate of $0.1$. The fine-tuning is performed for 2 epochs in case of ICHI and 10 epochs for CADEC. For the baseline models, we use the default hyperparameter values.

\subsection{Performance Evaluation}\label{sec:perf-eval}

We evaluate the classification models in terms of accuracy and F1-score (preferred more in case of class-imbalanced data like CADEC). 

\begin{table}[!ht]
\centering
\caption{Performance comparison of \medbert\ and baseline models. * indicates second best-performing model. 
 }
\label{tab:ichi-perf-compare}
\scalebox{0.73}{
\begin{tabular}{p{1.8cm}|p{0.6cm}|p{0.5cm}p{0.5cm}p{0.5cm}p{0.5cm}p{0.5cm}p{0.5cm}p{0.5cm}p{0.5cm}p{0.5cm}p{0.5cm}} \hline
& ICHI &\multicolumn{9}{c}{CADEC}\\
& ALL &\multicolumn{2}{c}{UPD}&\multicolumn{2}{c}{MAS}&\multicolumn{2}{c}{DM}&\multicolumn{2}{c}{IS}& \multicolumn{2}{c}{ALL} \\
\textbf{Models}&\textbf{Acc} &\textbf{Acc} &\textbf{F1}&\textbf{Acc} &\textbf{F1}&\textbf{Acc} &\textbf{F1}&\textbf{Acc} &\textbf{F1}&\textbf{Acc}&\textbf{F1}\\ \hline
ICHI 2016 Challenge Winners~\cite{verma2016classification} & 0.68 &-&-&-&-&-&-&-&-&-&-\\
SoA-DN~\cite{jalan2018medical} & 0.6&-&-&-&-&-&-&-&-&-&- \\ 
H-BLSTM + TFIDF-DN + SoA-DN~\cite{jalan2018medical} & \textbf{0.7}&-&-&-&-&-&-&-&-&-&- \\ 
TFIDF + SVM & 0.64 &0.82&0.59*&0.83&0.7&0.91&0.74&0.92&0.58&0.87&0.65\\
FastText &0.61&0.86&0.46&0.81&0.45&0.88&0.47&0.82&0.48&0.84&0.47 \\
TextCNN &0.58&0.83&0.47&0.87&0.72&0.9&0.69&0.92&0.52&0.88&0.6 \\
HAN &0.61& 0.83& \textbf{0.6}&0.82&0.6&0.88&0.61&0.92&0.48&0.86&0.57 \\
LCF-BERT & 0.69* &0.86&0.46&0.81&0.45&0.88&0.47&0.92&0.48&0.87&0.47\\ \hline
\medbert &  \textbf{0.7}&0.84&0.55&0.88&{\bf 0.74}&0.92&0.8*&0.95&{\bf 0.75}&\textbf{0.9}&\textbf{0.71} \\ \hline
Bert-base &0.65&0.84&0.32&0.86&0.72&0.91&0.52&0.94&0.54&0.89&0.53 \\
\medbert\ (Global only) & 0.68&0.79&0.55&0.86&\textbf{0.74}&0.93&\textbf{0.81}&0.94&0.73&0.88&\textbf{0.71}\\
\medbert\ (Local only) & 0.68&0.78&0.55&0.87&0.72&0.91&0.77&0.94&0.74*&0.88&0.7\\
 \hline
\end{tabular}}
\end{table}

\noindent \textbf{Discussion of Results.} 
 Our proposed model, \medbert\ achieves the highest performance in terms of both ICHI and CADEC dataset, which is shown in Table~\ref{tab:ichi-perf-compare}. We observe that the Bert-base model is able to outperform the state-of-the-art model~\cite{jalan2018medical} for the ICHI data, without using any medical features. We perform comparably with the 3-component ensemble model of Jalan et al.~\cite{jalan2018medical} comprising of TF-IDF, medical domain features like SoA-DN (second row of Table~\ref{tab:ichi-perf-compare}), and a neural network model (Hierarchical BiLSTM or HAN). However, we outperform their standalone `SoA-DN' model by $16.7\%$. The performance difference between \medbert\ and variants versus the competing baseline models is statistically significant in terms of McNemar's test, for all models except LCF-BERT for ICHI data.
 However, on CADEC dataset, we observe that the \medbert\ outperforms LCF-BERT by a significant margin (average F1 score of $0.71$ v/s $0.47$). We also notice that the classification accuracy of all models improved dramatically (average model accuracy increased from $0.65$ to $0.84$) on the multi-label dataset of CADEC as compared to ICHI data ($11.3\%$ of training data have more than one label as positive), implying that MFQC is better suited as a multi-label classification task. 

 \noindent \textbf{Ablation Analysis.}
 For ICHI data, \medbert\ significantly outperforms \medbert\ (Global only) and \medbert\ (Local only) models by $2.9\%$, thus highlighting the importance of the dual encoder architecture in our case. However, in case of CADEC, \medbert\ along with Global only and Local only variants, perform comparably. This may be attributed to the fact that the medical concepts filtered by the medical aspect extraction module of \medbert, do not completely accommodate CADEC information need categories that are focused on a sub-domain of adverse drug events only. Next, we observe that \medbert\ outperforms Bert-base (may be thought of as \medbert\ w/o medical keywords) by $7.7\%$, and shows the importance of adding medical side-information to these pretrained models.
 
\begin{figure}[!ht]
\centering
\includegraphics[scale=0.6]{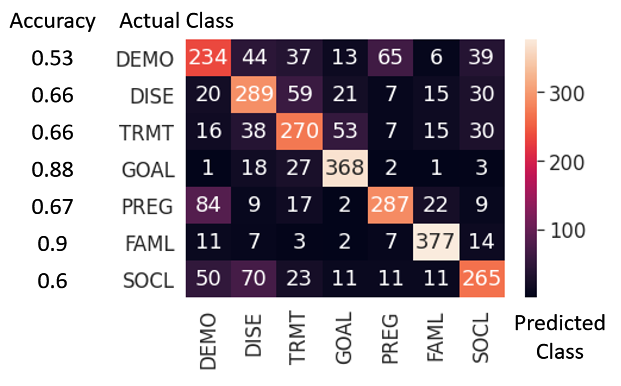}
\caption{Confusion matrix of \medbert\ on ICHI data along with class-wise accuracy scores, highlighting the most frequent misclassifications: PREG $\rightarrow$ DEMO, DEMO $\rightarrow$ PREG, and SOCL $\rightarrow$ DISE }
\label{fig:medbert-error}
\end{figure}

\noindent \textbf{Error Analysis on ICHI.} From Figure~\ref{fig:medbert-error}, we observe that \medbert\ fails most frequently in the following cases (true $\rightarrow$ predicted): PREG $\rightarrow$ DEMO, DEMO $\rightarrow$ PREG, and SOCL $\rightarrow$ DISE. On manual inspection and even from definition of both classes, we observe that `PREG' class targets a specific demographic, both age and gender, where topics like difficulties with conception, mother and unborn child's health during pregnancy, are covered. Thus, by definition, all data points labeled as `PREG', should also hold true for `DEMO'. Contrary to other target classes, Socializing (SOCL) do not target any health-related issues and rather focuses on recreational activities or hobbies. Thus, forcing \medbert\ to focus on medical concept-bearing words does not contribute towards improving the classifier performance of \textit{SOCL} class. 

\begin{figure}[!ht]
\centering
\includegraphics[scale=0.6]{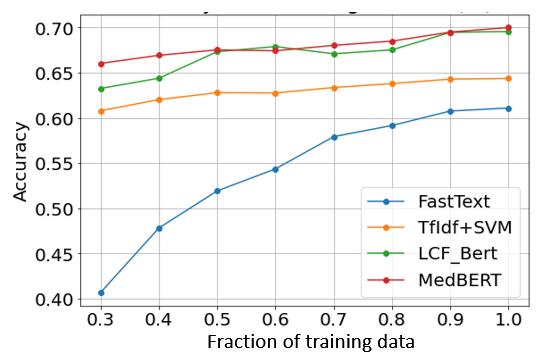}
\caption{Performance comparison of \medbert\ and baselines for different fractions of training data on ICHI data}
\label{fig:ichi-training-percent}
\end{figure}

\noindent \textbf{Effect of Training Data Size.} To evaluate the generalization power of different models, we evaluate them on different training data sizes of ICHI dataset (see Figure~\ref{fig:ichi-training-percent}). First, we observe that 
LCF-BERT and \medbert\ outperform other methods by a  large margin. For small train-sets, these approaches leverage the knowledge from pretrained models and use the training data only for fine-tuning the representations. Further \medbert\ outperforms LCF-Bert which showcase the advantage of using domain specific additional side information in the form of medical terms or aspects. Specifically for 30\% training data, \medbert\ outperforms LCF-BERT by $4.76\%$.

\section{Conclusion}\label{sec:conclusion}
We propose \medbert, a novel application of transformers-based dual encoder model, for MFQC task, which is also medical domain knowledge-aware. We contribute a multi-label MFQC dataset; \medbert\ achieves state-of-the-art performance on ICHI~(accuracy of $0.7$) and CADEC dataset (accuracy of $0.9$ and macro F1 score of $0.71$), and generalizes very well in low-resource settings. Through extensive experimentation, we learn that incorporating medical concept-bearing terms as side information, contribute significantly to \medbert. We learn that certain target classes heavily depend on \textbf{keywords}, while others require one to learn optimal representation of \textbf{medical context}. An interesting future direction will be to extend \medbert\ to structured prediction tasks like entity and relation prediction, or broadly link prediction. Instead of BERT, we will work with BioBERT~\cite{biobert2019}, which is a domain-specific pretrained model trained on biomedical articles.

\noindent \textbf{Acknowledgements.} This work is supported in part by the Institute PhD Fellowship of IIT Kharagpur, the Federal Ministry of Education and Research (BMBF), Germany under the project LeibnizKILabor (grant no. 01DD20003), the Adobe-funded project titled ``Computational Aspects and Role of Content for Persuasive Brand Positioning'', and IMPRINT-1 Project RCO (project no. 6537).

\bibliographystyle{ACM-Reference-Format}
\balance
\bibliography{sample-sigconf}


\end{document}